# Low Cost Eye Tracking for Vision Screening

Maissa Abir Smaili
*Istanbul Medipol University*
maissa.smaili@std.medipol.edu.tr

Yaman Ismaiel
*Istanbul Medipol University*
yaman.ismaiel1@std.medipol.edu.tr

Zafer Işcan
*Istanbul Medipol University*
zafer.iscan@medipol.edu.tr

**Abstract**—This engineering study assesses gaze-assisted data acquisition during visual screening with a Gaze Quest implementation via webcam and GC308 near infrared camera with an Orlosky eye tracking pipeline. Ten subjects performed under both acquisition conditions. Accuracy of Tumbling E orientation, reported application logMAR value, and response latency were determined from keyboard responses and hence are behavioral outputs as opposed to eye tracker screening results. Frame-to-frame change in the angle between successive normalized gaze vectors (28.2 degrees) served as the GC308 eye tracker derived diagnostic with an effective sampling rate of roughly 8 FPS. No screen target calibration or gaze to screen translation was done on the GC308 eye tracker output data for the current analysis and hence, these data represent acquisition stability and not gaze accuracy in relation to the target. The webcam pipeline relied on 25-point screen calibration. Because the display resolution and pixel pitch could not be retained in the records of the study, the reported application logMAR values cannot be viewed as measures of visual acuity in any way.

## I. Introduction and Related Work

For vision testing, feedback from the user is required. It is difficult when there are users who cannot provide feedback verbally or physically. In such situations, eye tracking can be used to gauge the individual's perception or fixation of some visual stimuli.

In this project, a prototype has been created which tracks eye movements during visual tasks performed by the user. For the prototype, the initial approach was to use the web camera along with the WebGazer framework [1], but the final prototype used a customized python pipeline with the help of OpenCV [2] and MediaPipe [3]. There have been recent literature reviews showing that there is an increased use of consumer cameras in eye tracking research [4].

In this current research, there are four areas which have been highlighted for the research: consumer camera gaze acquisition, vision screening task presentation, metric extraction, and descriptive analysis of outputs of webcam and GC308 IR-camera. Since both trackers give out inconsistent gaze metrics, the major research question will be that of whether it is possible to gather synchronized data using the prototype during vision screening tasks.

Eye tracking via the Internet has been considered as one of the promising alternatives to laboratory eye tracking. It was shown that eye tracking can be carried out with the use of a common webcam and interaction data of users [1]. Inexpensive eye-tracking techniques and gaze estimation systems in consumer devices have been analyzed [5], [6]. However, despite the advantages of these approaches, they still depend on certain limitations regarding their settings and precision. Recent studies revealed that under the controlled conditions, webcam eye tracking can compete with laboratory eye tracking [7], and the gaze estimation with the help of consumer devices using deep learning became more precise [8], [9].

Calibration is one of the important issues in eye-tracking studies. According to previous literature, the following may impact on the accuracy of gaze prediction: calibration quality [5], [10], image quality and illumination [8], [11], head position and eye localization [12], as well as more general camera and platform parameters [6]. Issues concerning the robustness to the variation in head-pose and image-quality have also been discussed for deep gaze estimation [13]. Standard eye-tracking methodology stresses the importance of viewing geometry, calibration, and individual differences [14].

Other applications involve the testing of acuity and ocular motor skills. Automated vision testing in children may be conducted based on the evaluation of eye movements in reaction to visual stimuli [15]. Smooth pursuit [16] and fixations [17] .were investigated in clinical and perception studies. The current investigation draws its inspiration from this research by integrating the acuity, contrast sensitivity, fixation, saccade, and smooth pursuit into one prototype. In case of the external camera experiment, the Orlosky 3D Eye Tracking Pipeline [18] is chosen to be the benchmarking software architecture as it computes the position of the eye-sphere center and the gaze vector for each infrared eye image using model-based geometry approach discussed in Hansen and Ji [19].

## II. System Design and Methods

The system architecture consists of input cameras, gaze estimator, calibration and validation process, visual test, metrics computation, and exporting data to a CSV file. The gaze estimation is done through landmarks of the face and iris using OpenCV [2] and MediaPipe [3].. Fig. 1 describes the test algorithm adopted for use in this project.

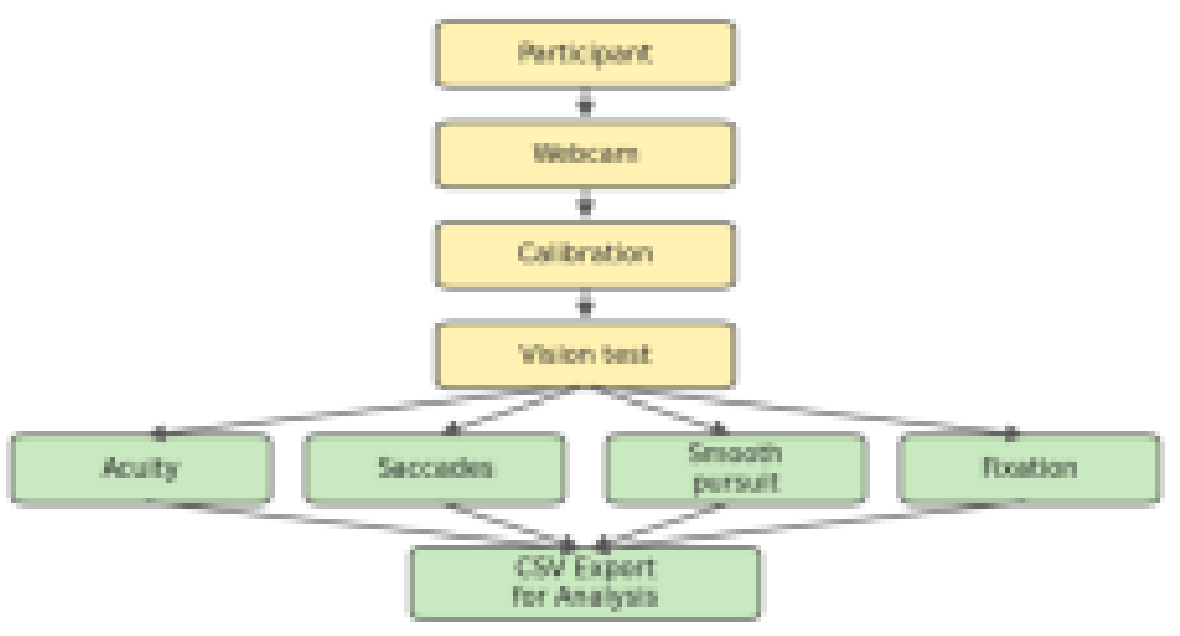

Fig. 1.workflow of the gaze based vision screening prototype.

The visual tests consist of acuity, contrast sensitivity, saccades, smooth pursuit, and fixation. These tests were considered to be included because of their ability to measure target detection, gaze shifting, target tracking, and gaze stability. Once each session ends, the system outputs the test results in the form of trial and gaze sample points. The GC308 infrared camera was analyzed with respect to its eye-tracking accuracy via the open-source Orlosky eye tracking pipeline [18]. This pipeline uses model-based estimation to calculate the eye-sphere center and the gaze vector per frame [19]. The subject-level testing protocol was kept the same in both conditions. Whereas Gaze Quest employed a screen-target calibration procedure along with a gaze to screen mapping, neither one of these was used in the case of GC308 / Orlosky in this particular analysis. This means that any calibration error or screen coordinate accuracy could not be determined in this case. The output was therefore measured only for diagnostic purposes as described above.

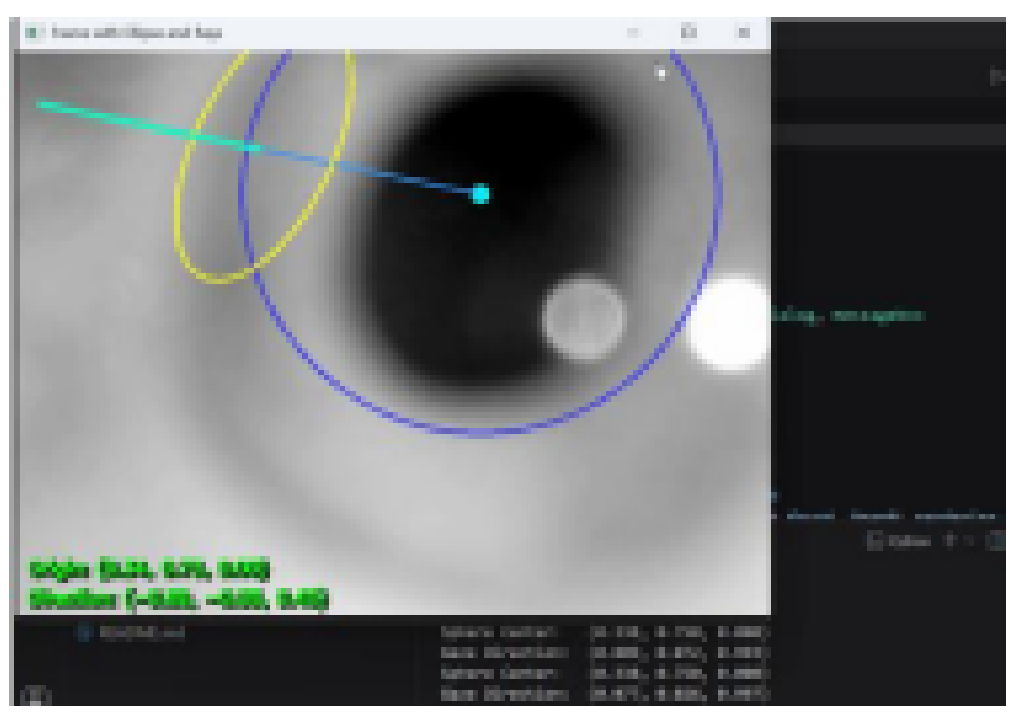

Fig. 2. Example GC308/Orlosky infrared eye frame with eye-sphere and gaze vector overlays.

### *Visual Tasks and Data Logging*

The Gaze Quest test involved calibration and validation of a 25-point screen. The acuity/contrast module showed a Tumbling-E optotype and saved 22 high-contrast and 22 low-contrast trials during each session. The software used the entered 14-inch screen diagonal length and a viewing distance of 50 cm to compute the size of the optotypes in terms of nominal logMAR. Responses for orientation were inputted using the keyboard or by time-out; thus, response accuracy and latency become behavioral measures in this task rather than eye-tracking measures. The display resolution and pixel pitch and hence the minimum renderable optotype size in arcminutes and the minimum valid logMAR measure were not recorded in this experiment. Therefore, the reported logMAR measures can be considered as only the stimulus levels reported by the application but not as the visual acuity measures.

The saccade module provided data from 20 trials that were equally distributed among eight target directions. Smooth pursuit included eight blocks with horizontal, vertical, circular, and figure-eight eye movements at two different speeds and one fixation block for assessing eye movement stability. These tests resulted in 73 trials without practice per session. The target eccentricities and timings that are required for saccade characterization in a clinical setting are missing in this dataset. However, more importantly, the GC308/Orlosky system has been demonstrated to work effectively with a rate of 8 FPS. Consequently, parameters such as saccadic onset, duration, latency, peak velocity, pursuit gain, etc., which require high frequency eye movement recordings, have not been calculated.

## III. Experimental Protocol

Ten participants completed both conditions under controlled indoor lighting while viewing a 14 inch display from approximately 50 cm. Raw session timestamps show a fixed sequence: webcam/Gaze Quest first and GC308/Orlosky second, typically within the same visit. Each condition contained 730 records, yielding 1,460 records overall. The fixed order is acknowledged as a possible practice or fatigue confound. The study protocol was approved by the Istanbul Medipol University Non-Interventional Clinical Research Ethics Committee (Decision No. 491, 26 March 2026). The approved submission included the research protocol and informed volunteer consent documentation. The present manuscript reports de-identified participant-level summaries.

### *Outcome Measures and Statistical Analysis*

Accuracy of participant-level Tumbling-E orientation-response task was evaluated using the results of the 44 acuity contrast trials performed per session. Participant-level mean application-reported logMAR and response time measures were also provided. All three measures were obtained using the results of the task presentation and keyboard-based responses and, hence, were behavioral/application-level output measures and not the GC308 screening results. The paired two-tailed t-test was used since the same 10 participants took part in both conditions. Also, the Wilcoxon signed rank test was utilized for response accuracy as a sensitivity analysis. A p-value greater than 0.05 indicates a lack of difference, not equivalence. The diagnostics from eye tracker were evaluated independently. CSV files from webcam yield an error in terms of screen coordinates, while GC308 and Orlosky export a 3D gaze vector. Gaze-vector stability was the main diagnostic from the GC308 eye tracker used in the study, which was measured as the angle between two consecutive gaze vectors; its mean value was 28.2°. Effective sampling rate and automatic rejection categories were also calculated. Since there was no screen-target mapping in GC308, these diagnostics do not form any common measure of gaze accuracy.

## IV. Results

The behavioral data were consistent between the two consecutive experiments. Keyboard responses yielded accuracies of 91.4% and 93.2% in the webcam condition and the GC308 condition, respectively; this did not reach statistical significance between conditions. Similar results were obtained in application-measured logMAR and response latency, as can be seen in Table I and Fig. 3. The behavioral data do not reflect any outcomes of the GC308 test screen and the absence of a difference does not mean that the two procedures are equally effective. These results simply indicate that the prototype could produce consistent behavioral recordings in both conditions.

TABLE I. Behavioral Metrics by Tracking Condition

| Metric | Webcam | GC308 | p |
|---|---|---|---|
| Accuracy (%) | 91.4 ± 18.0 | 93.2 ± 11.8 | 0.57 |
| logMAR acuity | 0.97 | 1.00 | 0.34 |
| Latency (ms) | 595 | 600 | 0.56 |

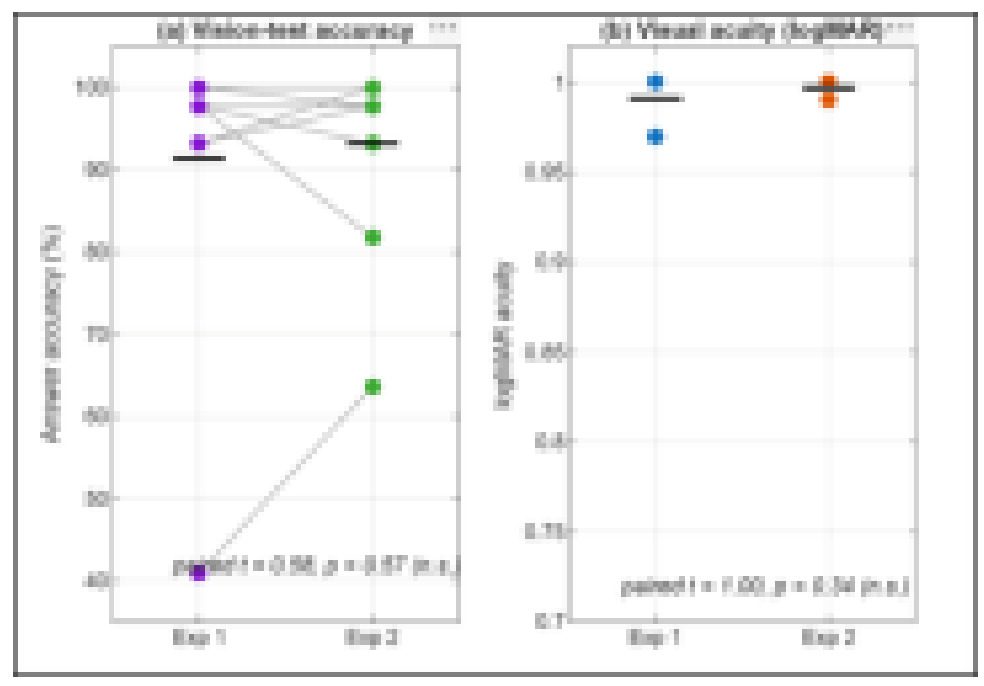

Fig. 3. Behavioral results by tracking technique for answer accuracy and logMAR visual acuity.

Accuracy at the user level indicated that most participants performed well in both conditions. Nine out of ten participants had an accuracy of at least 80% in both conditions, while only one of them had a low level of accuracy. Fig. 3 shows the accuracy distribution for users.

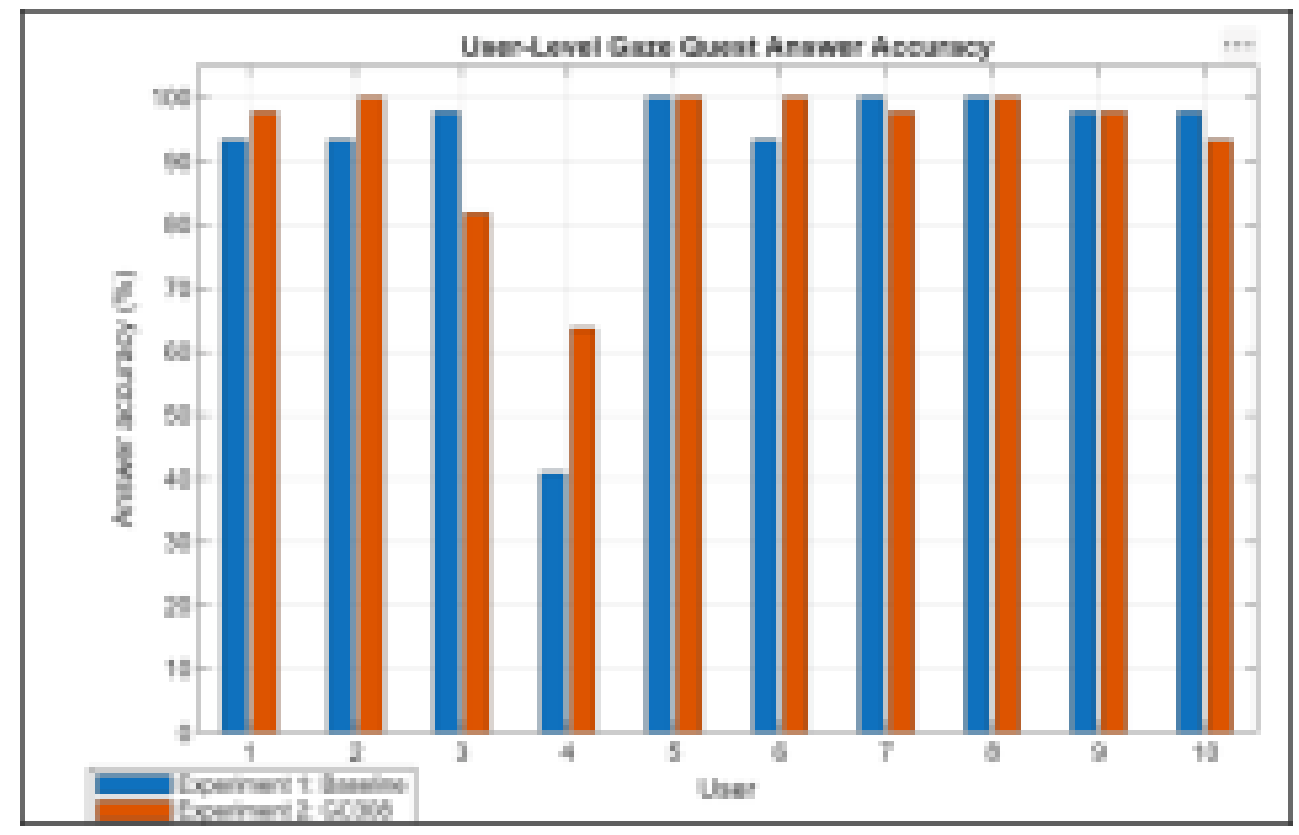

Fig. 4. User-level answer accuracy for the webcam baseline and GC308 conditions.

The difference between the two types of gaze signal diagnostics was not only in scale but also in meaning. While the webcam outputs generated an error diagnostic based on the screen coordinates, with the average of 10.5 degrees, the GC308/Orlosky pipeline generated a gaze vector angular step diagnostic, with the average of 28.2 degrees. It is an outcome of eye tracker calibration and not the screening or accuracy measure for target coordinates. Since the two diagnostics have different scales, their values cannot be used to rate gaze estimation accuracy. Thus, Fig. 5 presents the results descriptively.

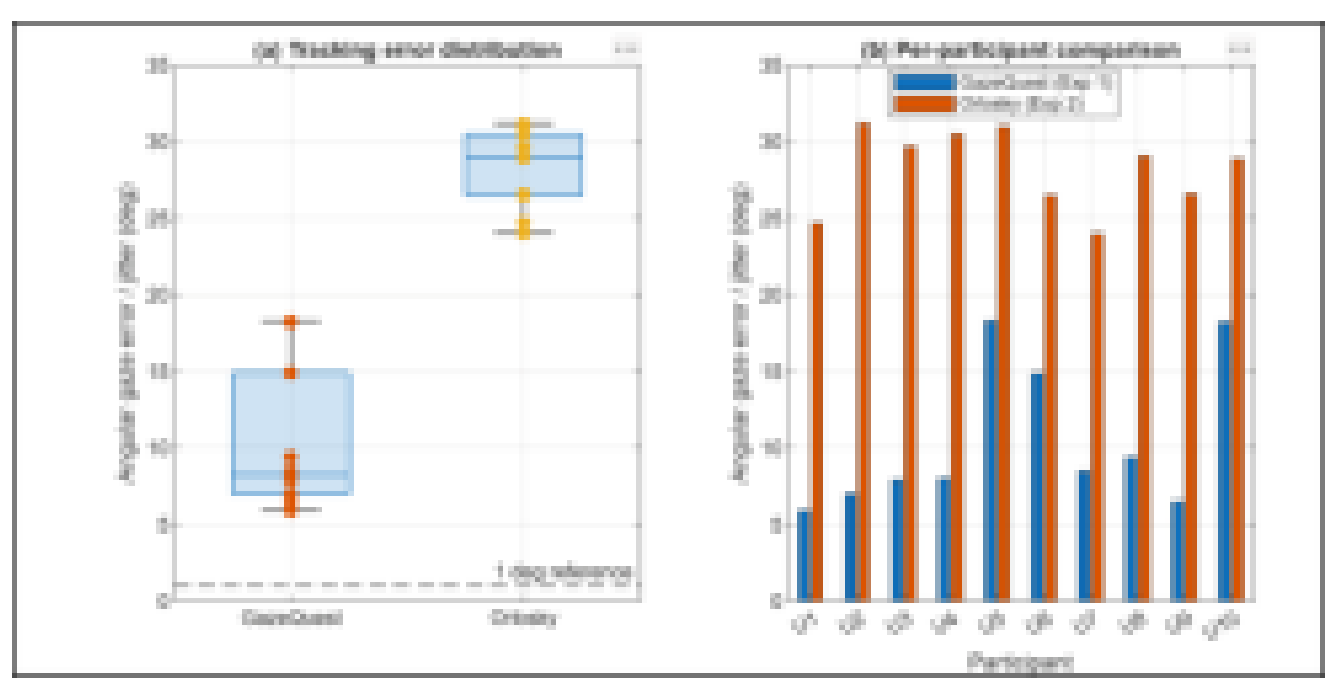

Fig. 5. Gaze angle error by tracking modality and participant.

The average effective sampling rate of the GC308/Orlosky system was roughly 8 FPS, with significant variability across subjects (Fig. 6). This is an adequate rate to illustrate the possibility of acquiring the gaze vectors on a frame by frame basis but is still highly undersampled for capturing any rapid eye movements. Hence, there were no kinematic or clinical measures from saccades or smooth pursuits reported here.

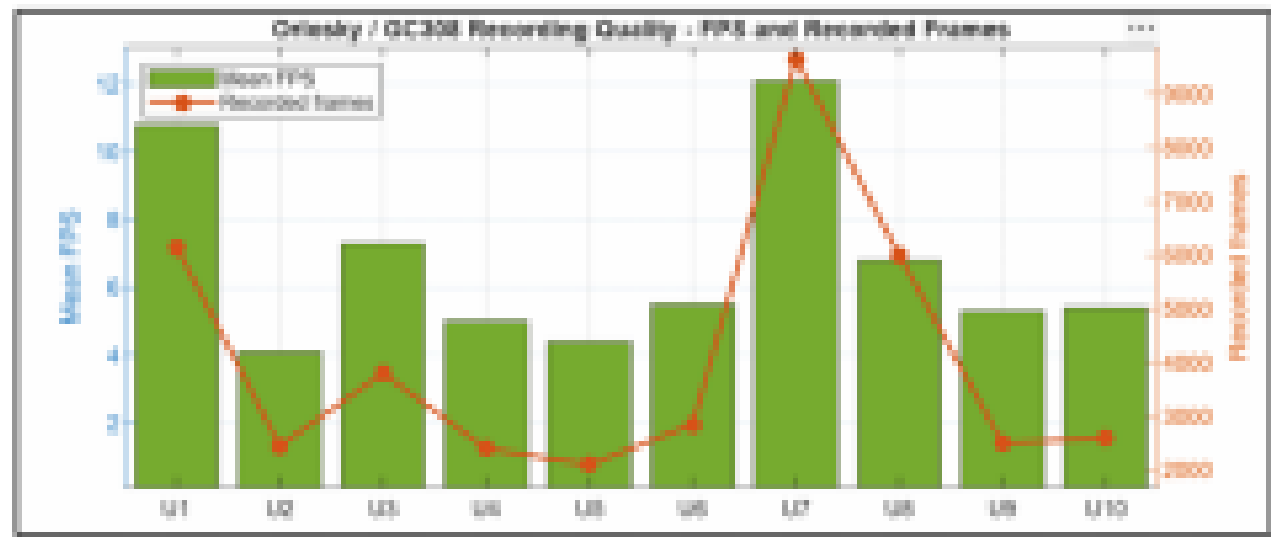

Fig. 6. GC308 and Orlosky recording quality showing mean sampling rate and recorded frames.

Autocriteria-based rejection revealed the primary failure mode for each condition. For the webcam condition, calibration error was the predominant factor causing rejection, while for the infrared condition, low valid sample ratio and fixation timeout were responsible. The factors behind rejections are listed in Fig. 7 below.

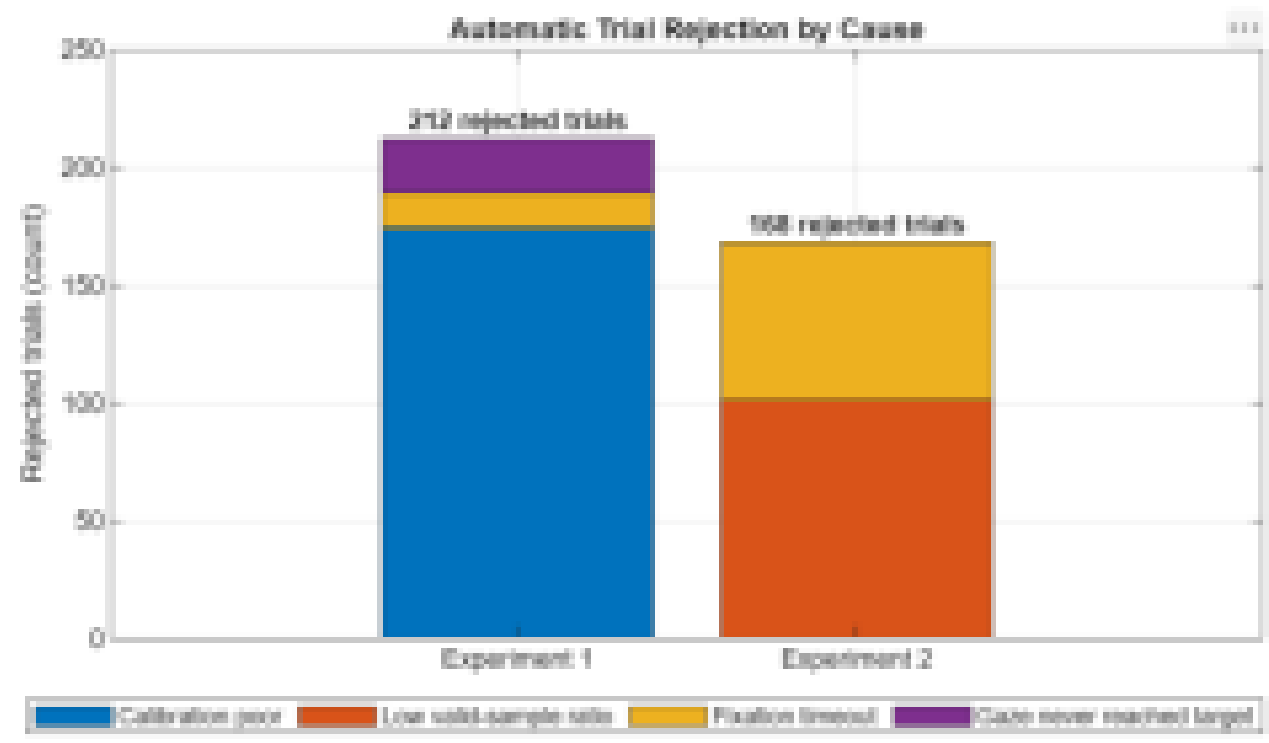

Fig. 7. Automatic trial rejection by cause across the two experimental conditions.

## V. Discussion

The Technical findings far outweigh clinical: synchronization of task records and gaze data was successfully achieved by all ten participants through consumer-level camera hardware. Behavioral Tumbling-E accuracy, nominal logMAR value, and response latency were measured based on keyboard responses and thus cannot be considered results from GC308 screening. Rather, GC308's contribution here can be seen in three-dimensional gaze vector acquisition and the stability/recording diagnostic measurements obtained from it. Thus, it is most appropriate to understand this project as a gaze data collection pilot performed while performing vision screening tasks, but not validation of clinical screening instrument itself.

There are two fundamental differences between gaze pipelines in question. First of all, while Gaze Quest relied on 25 screen targets, current GC308/Orlosky implementation does not translate camera coordinate system gaze vectors into any particular screen targets. Thus, target referenced accuracy and calibration quality cannot be estimated retrospectively for GC308 case. Moreover, with approximately 8 FPS effective rate it is impossible to estimate rapidly occurring saccades and analyze smooth pursuit kinematics. Future comparison should be performed with matching display targets, gaze to screen mapping, angular precision/accuracy definitions and considerably higher acquisition rate.

Care must be taken regarding nominal logMAR. The software employed a diagonal of 14 inches and 50 cm for viewing, but neither resolution nor pixel pitch was saved, so it is impossible to reconstruct the smallest readable letter and measurement scale. Numbers above 0.97-1.00 are stimulus values reported by the software, not true acuities. The display specifications will be recorded in future work. Neither bill of materials nor cost comparison was included; cost-effective claims have been removed.

## VI. Conclusion

The current pilot experiment shows synchronization of recording of behavior task performance and gaze while conducting a vision test task by means of webcam-based pipeline and GC308/Orlosky 3D gaze vector tracking. Tumbling-E performance accuracy, nominal logMAR, and response latencies are behavioral/application-level outputs rather than eye tracker testing outcomes. In the case of GC308, the eye tracker outcome is a descriptive gaze vector stability diagnostics (frame-to-frame mean angular step is 28.2 degrees) with an effective sample rate of about 8 FPS. Because the GC308 eye tracker was not calibrated for screen targets, the current data does not allow establishing target-based gaze vector accuracy. Furthermore, because of the low sample rate, neither saccades nor smooth pursuit can be measured reliably. Pixel pitch on the screen display was lost during calibration, thus, the nominal logMAR cannot be interpreted clinically. The current results thus show only engineering feasibility and data collection reproducibility.

## References

[1] A. Papoutsaki, P. Sangkloy, J. Laskey, N. Daskalova, J. Huang, and J. Hays, "WebGazer: Scalable webcam eye tracking using user interactions," in Proc. 25th Int. Joint Conf. Artif. Intell. (IJCAI), 2016, pp. 3839–3845.

[2] G. Bradski, "The OpenCV library," Dr. Dobb's J. Softw. Tools, vol. 25, no. 11, pp. 120–125, Nov. 2000.

[3] C. Lugaresi et al., "MediaPipe: A framework for building perception pipelines," 2019, arXiv:1906.08172.

[4] I. Rakhmatulin, "A review of the low-cost eye-tracking systems for 2010–2020," 2020, arXiv:2010.05480.

[5] G. Garde, A. Larumbe-Bergera, B. Bossavit, S. Porta, R. Cabeza, and A. Villanueva, "Low-cost eye tracking calibration: A knowledge based study," Sensors, vol. 21, no. 15, Art. no. 5109, Jul. 2021, doi: 10.3390/s21155109.

[6] A. Kar and P. Corcoran, "A review and analysis of eye-gaze estimation systems, algorithms and performance evaluation methods in consumer platforms," IEEE Access, vol. 5, pp. 16495–16519, Aug. 2017, doi: 10.1109/ACCESS.2017.2735633.

[7] T. Kaduk, C. Goeke, H. Finger, and P. König, "Webcam eye tracking close to laboratory standards: Comparing a new webcam based system and the EyeLink 1000," Behav. Res. Methods, vol. 56, no. 5, pp. 5002–5022, Aug. 2024, doi: 10.3758/s13428-023-02237-8.

[8] S. Saxena, L. K. Fink, and E. B. Lange, "Deep learning models for webcam eye tracking in online experiments," Behav. Res. Methods, vol. 56, no. 4, pp. 3487–3503, Jun. 2024, doi: 10.3758/s13428-023- 02190-6.

[9] K. Krafka, A. Khosla, P. Kellnhofer, H. Kannan, S. Bhandarkar, W. Matusik, and A. Torralba, "Eye tracking for everyone," in Proc. IEEE Conf. Comput. Vis. Pattern Recognit. (CVPR), 2016, pp. 2176– 2184, doi: 10.1109/CVPR.2016.239.

[10] P. Blignaut, "Mapping the pupil-glint vector to gaze coordinates in a simple video-based eye tracker," J. Eye Mov. Res., vol. 7, no. 1, Art. no. 4, pp. 1–11, 2014, doi: 10.16910/jemr.7.1.4.

[11] X. Zhang, Y. Sugano, M. Fritz, and A. Bulling, "Appearance based gaze estimation in the wild," in Proc. IEEE Conf. Comput. Vis.
Pattern Recognit. (CVPR), 2015, pp. 4511–4520, doi: 10.1109/CVPR.2015.7299081.

[12] R. Valenti, N. Sebe, and T. Gevers, "Combining head pose and eye location information for gaze estimation," IEEE Trans. Image Process., vol. 21, no. 2, pp. 802–815, Feb. 2012, doi: 10.1109/TIP.2011.2162740.

[13] S. Park, A. Spurr, and O. Hilliges, "Deep pictorial gaze estimation," in Computer Vision – ECCV 2018, Lecture Notes in Computer Science, vol. 11217. Cham, Switzerland: Springer, 2018, pp. 741–757, doi: 10.1007/978-3-030-01261-8_44.

[14] K. Holmqvist, M. Nyström, R. Andersson, R. Dewhurst, H. Jarodzka, and J. van de Weijer, Eye Tracking: A Comprehensive Guide to Methods and Measures. Oxford, U.K.: Oxford Univ. Press, 2011.

[15] J. J. M. Pel, J. C. W. Manders, and J. van der Steen, "Assessment of visual orienting behaviour in young children using remote eye tracking: Methodology and reliability," J. Neurosci. Methods, vol. 189, no. 2, pp. 252–256, Jun. 2010, doi: 10.1016/j.jneumeth.2010.04.005.

[16] R. J. Leigh and D. S. Zee, The Neurology of Eye Movements, 5th ed. New York, NY, USA: Oxford Univ. Press, 2015, doi: 10.1093/med/9780199969289.001.0001.

[17] S. Martinez-Conde, S. L. Macknik, and D. H. Hubel, "The role of fixational eye movements in visual perception," Nat. Rev. Neurosci., vol. 5, no. 3, pp. 229–240, Mar. 2004, doi: 10.1038/nrn1348.

[18] J. Orlosky, "EyeTracker: A lightweight, robust 3D eye tracker in Python," GitHub repository, 2024. [Online]. Available: https://github.com/JEOresearch/EyeTracker. [Accessed: Aug. 11, 2026].

[19] D. W. Hansen and Q. Ji, "In the eye of the beholder: A survey of models for eyes and gaze," IEEE Trans. Pattern Anal. Mach. Intell., vol. 32, no. 3, pp. 478–500, Mar. 2010, doi: 10.1109/TPAMI.2009.30.